\documentclass{article}

\usepackage{graphicx}
\usepackage{makecell}
\usepackage{authblk}
\usepackage[T1]{fontenc}    
\usepackage{hyperref}       
\usepackage{url}            
\usepackage{booktabs}       
\usepackage{amsfonts}       
\usepackage{nicefrac}       
\usepackage{microtype}      
\usepackage{xcolor}         
\newcommand{\edit}[1]{{#1}}
\begin{document}
\date{}

\title{VR.net: A Real-world Dataset for Virtual Reality Motion Sickness Research}
\author[1]{Elliott Wen}
\author[2]{Chitralekha Gupta}
\author[2]{Prasanth Sasikumar}
\author[1]{Mark Billinghurst}
\author[3]{James Wilmott}
\author[3]{Emily Skow}
\author[3]{Arindam Dey}
\author[2]{Suranga Nanayakkara}
\affil[1]{The University of Auckland}
\affil[2]{National University of Singapore}
\affil[3]{Meta}

\maketitle


\begin{abstract}

Researchers have used machine learning approaches to identify motion sickness in VR experience. 
These approaches demand an accurately-labeled, real-world, and diverse dataset for high accuracy and generalizability.
As a starting point to address this need, we introduce `VR.net', a dataset offering approximately 12-hour gameplay videos from ten real-world games in 10 diverse genres. For each video frame, a rich set of motion sickness-related labels, such as camera/object movement, depth field, and motion flow, are accurately assigned.
Building such a dataset is challenging since manual labeling would require an infeasible amount of time. Instead, we utilize a tool \edit{to} automatically and precisely extract ground truth data from 3D engines' rendering pipelines without accessing VR games' source code.
We illustrate the utility of VR.net through several applications, such as risk factor detection and sickness level prediction.
We continuously expand VR.net and envision its next version offering 10X more data than the current form.
We believe that the scale, accuracy, and diversity of VR.net can offer unparalleled opportunities for VR motion sickness research and beyond.
\end{abstract}

\section{Introduction}
VR gaming has gained widespread popularity in recent years, with the annual market revenue projected to reach \$87 billion by 2030\footnote{https://bloom.bg/3INVQ9O}. 
However, up to 40\% of users suffer from VR motion sickness with symptoms like fatigue, disorientation, and nausea \cite{johnson2005introduction}. 
These adverse effects can severely undermine the user experience. 

To date, considerable research~\cite{saredakis2020factors,chang2020virtual,oh2022cybersickness} has uncovered that VR content is the most influential factor for VR sickness.
Therefore, many VR game stores desire to identify motion sickness risk factors and predict comfort levels in VR content. 
The results could inform users about potentially unpleasant feelings and assist game developers in adjusting the game-level design for a safer experience. 
Initially, this evaluation process was conducted by human experts, which quickly failed to cope with the growing number of VR games.
Recently, researchers have proposed using Machine Learning (ML) approaches to identify motion sickness's presence~\cite{oh2022cybersickness,hell2018machine}.
Despite progress, many of these studies pointed out that the limited training datasets constrain their models. 
To improve accuracy, they would require a large-scale dataset containing many hours of video clips and accurate risk factors~\cite{padmanaban2018towards}. The video clips should also come from diverse real-world game genres to ensure generalization. To our knowledge, such a comprehensive dataset does not exist.

In this paper, we present a new dataset called `VR.net'. It is publicly accessible via cloud storage\footnote{https://vrhook.ahlab.org} and possesses the following properties.
\begin{enumerate}
\item \textbf{Real-world, Diverse, and Large-scale:}
VR.Net aims to provide a diverse dataset for real-world VR games to stimulate  ML-based VR motion sickness research.
VR.net currently encompasses ten off-the-shelf VR games from 10 representative genres. Each game is evaluated by at least five different users, and each gameplay session lasts 15 minutes. This provided the first version of our dataset with around 12 hours of gameplay videos. We constantly expand VR.net and aim to offer 10X more data upon completion.

\begin{figure*}[t]
  \begin{center}
  \includegraphics[width=1\linewidth]{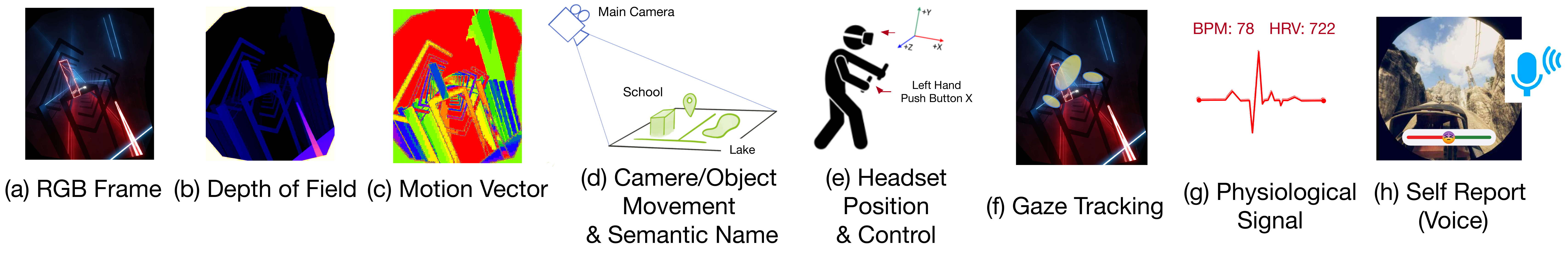}
  \caption{A snapshot of labels in VR.net.}
  \label{fig:labeldemo}
  \end{center}
  \end{figure*}

\item \textbf{Rich Labels:}
Each video is recorded at a frame rate of 30 frames per second.
For each frame, VR.net assigns 13 types of labels that are known to induce motion sickness. They are classified into two groups: `graphics' and `interaction'. The `graphics' labels describe game contents in each video frame, including camera location/speed/rotation, object location/speed/rotation, object semantic name, field of view, depth of field, and motion flow. 
The interaction labels describe a user's reaction toward each frame, such as headset movement, joystick movement, and self-report. 
The next version of VR.net also aims to exploit advanced VR headsets' sensors and provide gaze tracking and physiological signal.
Fig~\ref{fig:labeldemo} shows a snapshot of these labels. 

\item \textbf{Accuracy:}
VR.net aims to offer a clean dataset. Thus, we opt not to use error-prone manual or vision-based labeling.
Instead, we enhance \edit{our previous} data collection tool called VRhook~\cite{wen2022vrhook} to  automatically extract accurate ground-truth data from 3D game engines' graphics pipelines. This approach enables us to obtain high-quality labeled data without accessing VR games' source code.
\end{enumerate}

The rest of the paper is organized as follows. 
First, we summarize the differences between VR.net and other datasets in Section~\ref{related}.
We then demonstrate the construction process of VR.net and the implementation detail of our data collection tool in Section~\ref{construction}.
Section~\ref{sample} showcases several applications of the current VR.net, such as risk factor detection and comfort level prediction. We aim to show that VR.net can serve as a valuable resource for VR motion sickness research and broader machine-learning applications (e.g., object recognition and localization).

\section{VR.net and Related Datasets}~\label{related}
In this section, we compare VR.Net with other related datasets. We summarize their differences in Table~\ref{tab:comp-table}. 

\textbf{Manual Labeling:} 
Several manual-labeled datasets have served as training and evaluation benchmarks for ML-based VR motion sickness research.
For instance,
VRSA~\cite{kim2018vrsa} is acquired from the media-sharing platform Youtube. VRSA contains nine stereoscopic videos with beach, driving, and roller coaster scenes. Based on the scene movement speed, these videos are manually assigned three labels: `slow', `normal', and `intense'. 
VRSA has been utilized VRSA to demonstrate the correlation between exceptional movement patterns and VR sickness scores.
However, the dataset contains less than one-hour video content and thus is insufficient to train a complex machine learning model. 
In contrast, VR.Net aims to offer videos 100x the length of VRSA, allowing researchers to experiment with various state-of-the-art deep learning networks. 

\textbf{Vision-based Labeling:} Several researchers applied computer vision models to automate video labeling and generate a larger-scale dataset.
For instance, the study~\cite{padmanaban2018towards} assigned three labels (i.e., camera movement speed, direction, and depth) to 19 stereoscopic videos. 
These labels are approximated using principal component analysis on 2D pixel space.
Similarly, Lee et al.~\cite{lee2019motion} utilized the coarse-to-fine scale-invariant feature transform flow matching and the semi-global block matching to estimate motion flow and depth of field.
Due to the heuristic nature of these vision algorithms, the generated label sets are now prone to data noise, eventually decreasing ML models' performance.
In contrast, VR.net retrieves accurate ground-truth data from a VR game's rendering pipeline. No noise is induced in this process.

\textbf{Synthesized Datasets:} 
Several datasets are gathered from purposely-built VR simulation environments (i.e., synthesized)~\cite{hell2018machine,martirosov2022cyber,oh2022cybersickness,kim2018virtual}.
These environments usually provide users with passive VR experiences, such as a roller coaster or a space simulator. During a play-through, the environments can capture comprehensive game scene parameters for automatic labeling.
Though these synthesized datasets are accurate and more extensive in size, their data distribution might not reflect real-world VR games, which results in model generalization issues. 
In contrast, VR.net obtains labeled data from off-the-shelf VR games in 10 representative genres, covering most real-world VR experiences.

\begin{table}[t]
\centering
\caption{Comparison between VR.net and existing datasets}
\label{tab:comp-table}
\resizebox{0.9\textwidth}{!}{%
\begin{tabular}{|l|l|l|l|l|}
\hline
 & VR.Net & Manual Labelled~\cite{kim2018vrsa} & Vision-based~\cite{padmanaban2018towards} & Synthetic~\cite{hell2018machine} \\ \hline
Dataset Size & \makecell[l]{~12 Hours \\ (100 hours \\ in next version)}   & $<$2 Hours  & $<$2 Hours & $<5$ Hours \\ \hline
Label Quality & Precise & Prone to Noise & Prone to Noise & Precise \\ \hline
Label Types & Rich (13) & Limited (3) & Limited (5) & Rich (11) \\ \hline
Diversity  & \makecell[l]{Yes \\ (10 real-world genres)} & No (Passive videos) & No (Passive videos) & No (1 game) \\ \hline
\end{tabular}%
}
\end{table}

It should be noted that some datasets~\cite{liao2020using,smyth2021exploring,xue2021ceap} aim to reveal the correlations between users' physical responses and sickness levels. However, they provide very limited content-related labels. Thus, they cannot be used to train a motion sickness detector for VR content.

\section{Constructing VR.net}~\label{construction}
In this section, we describe our methodology for constructing VR.net.

\subsection{Collecting Candidate VR Games}
The first stage of VR.net construction is to select a vast number of real-world VR games from representative genres.
To accomplish this target, we crawl and analyze content from the popular online game store Steam\footnote{https://store.steampowered.com/}.
Though several other stores, such as Oculus Quest\footnote{https://www.oculus.com/experiences/quest/} or Epic\footnote{https://store.epicgames.com/en-US/} are also fast expanding, Steam still provides the most extensive collection of VR games. 

In detail, we first retrieve a complete title list of VR-only games using Steam's query API. Next, we utilize a third-party API named SteamSpy\footnote{https://steamspy.com/} to gather additional metadata such as game descriptions, categories set by developers, user tags, and estimated download numbers. 
It should be noted that the developer categories and user tags do not necessarily indicate the genre of games. Instead, they may describe a game's selling price (e.g., free to play), visual properties (e.g., breathtaking scenery), or program features (e.g., cross-platform). To compensate for this discrepancy, we invite three independent researchers to perform content analysis coding on the data. The coders follow the validated Lucas and Sherry genre classification system~\cite{foxman2021beyond}. 
They are allowed to assign multiple genres into a single game. 
To reduce the workload of the coders, we exclude the VR games with fewer than 1000 downloads from the dataset. They are usually early-access games and not for official release.
Note that these coders are experienced gamers and have previously received training on the coding protocols.
They are expected to categorize games more systematically and consistently.
We also conduct a spot-check on 5\% of the codes, and we do not discover any irregularities. 

Finally, the following genres are included in our dataset: 1) Action, 2) Adventure,
3) Fighter, 4) Flight, 5) Music, 6) Puzzle, 7) Racing, 8) Shooter,
9) Simulation, 10) Sports. 
For each genre, we conduct random sampling to select ten games.
The current VR.net constitutes 10\% of the candidate games and is constantly expanding.

\subsection{Dataset Label Selection}~\label{labellist}
The current version of VR.net offers approximately 12 hours of gameplay videos. 
For each video frame, VR.net automatically annotates a rich set of labels known to cause motion sickness. The labels are classified into `graphics' and `interaction'. The graphics group describes the game contents in each video frame:
\begin{enumerate} 
\item \textbf{Camera Movement:} 
In computer graphics, each frame is generated from a camera's perspective (in analogy to a real-life camera). 
Humans are more likely to experience nausea when the camera is moving than standing still~\cite{bonato2008vection, chardonnet2015visually, liu2012study}. 
Lo et al.~\cite{so2001effects,so2001metric} have quantitatively investigated the effects of fast camera translational movement speed on VR sickness. 
VR sickness can also arise from excessive camera accelerating or decelerating movements~\cite{keshavarz2015vection,kim2018virtual}. 
In addition, users can experience more significant discomfort when viewing a VR scene with rotational camera movements compared with translational movements~\cite{bonato2009combined}. VR.net records each camera's `view' matrix, which jointly encodes the camera location, velocity, acceleration, and rotation.

\item \textbf{Field of View:} The field of view (FOV) is the extent of the observable game world at any given moment. 
Many studies~\cite{duh2001effects,lin2002effects} have attributed the VR sickness to an overly wide FOV setting. Restricting the FOV can effectively relieve both subjective and objective symptoms of VR sickness \cite{adhanom2020effect}. VR.net captures each camera's `projection' matrix, which encodes FOV in a matrix form. 

\item \textbf{Depth of Field:}  Depth of Field (DoF or depth texture) measures the distance from a viewing camera to every pixel on the screen. 
Carnegie et al.~\cite{carnegie2015reducing} reported that inappropriate DOF settings could cause visual discomfort. 

\item \textbf{Motion Vector:} Motion vectors capture the per-pixel, screen-space motion of objects from one frame to the next. Several studies~\cite{lee2017estimating,park2022mixing} have revealed that intensive motion flows are one of the major factors that cause simulator sickness.

\item \textbf{Object Movement:}
The movement pattern of each object can also be a crucial factor in inducing motion sickness. A typical explanation is `vection', where stationary viewers feel like they have moved because nearby objects are approaching, receding, or rotating. Vection is believed to provoke motion sickness in susceptible individuals~\cite{keshavarz2015vection,keshavarz2019effect}.
VR.net captures the `model' transformation matrix for each visible object, which jointly encodes an object's location, velocity, acceleration, and rotation.
VR.net also provides each object's 3D bounding box, which can be used for object identification and collision detection.

\item \textbf{Object Semantic Name:}
Several studies~\cite{kim2020deep,nesbitt2017correlating} indicate that different VR game contents may provoke
varied levels and rates of nausea symptoms.
 In VR.net, we record each visible object's name for semantic modeling of VR scenarios. The names are initially assigned by content developers to facilitate object management. 
\end{enumerate}
The interaction group describes how a user reacts to a video frame:
\begin{enumerate}
\item \textbf{Headset Movement:}
Headset movement is tightly associated with the motion sickness level~\cite{li2021queasy,tychsen2020effects}. An excessive head movement may cause postural instability or vestibular-ocular reflex maladaptation, thus increasing the unpleasantness. VR.net retrieves the headset pose matrix from the underlying hardware, which accurately reflects a headset's position, orientation, and spatial velocity. 

\item \textbf{Joystick Control:} 
Several studies~\cite{dong2011control} have shown that a user tends to report more severe VR sickness if the game restricts the user's interactions with joysticks. Therefore, VR.net monitors each connected joystick's pose matrix and button actions.

\item \textbf{Self Report:} 
\edit{VR.net adapts a widely-used measuring protocol called Fast Motion sickness Scale (FMS)~\cite{keshavarz2011validating} to collect momentary self-reported comfort scores from users.
FMS asks the participant to verbally rate their experienced sickness every few minutes on a numerical scale (e.g., 1 to 5, where one indicates that the current scene does not incur motion sickness and five indicates that the scene results in severe discomfort). 
FMS is easy to administer and can collect measurements throughout a gameplay session. Verbal reports are also shown to be less distracting when compared to hand-action-based responses~\cite{proffitt1995perceiving}. }

\end{enumerate}
Apart from these labels, we are also investigating the possibility of collecting physiological and gaze-tracking signals, which are effective indicators of motion sickness severity~\cite{holmes2001correlation, diels2007visually}.
Several recently-released VR headsets (e.g., Meta Quest Pro)  provide built-in physiological sensors and gaze trackers. However, the data access APIs are still highly experimental. 
VR.net aims to include these two labels once the APIs are finalized.

\subsection{Data Collection Tool Implementation}
\edit{
In our previous study~\cite{wen2022vrhook}, we applied the dynamic hooking technique on Windows' low-level graphics stacks to harness ground-truth data from real-world VR games. The method does not demand access to a game's source code. 
In this study, we extended the previous method to access high-level graphics data from 3D game engines. 
The engine data has a well-documented format and can be easily interpreted by humans. As a result, it allows us to extract data in a highly-precise manner. It also enables us to capture more complex and semantic labels, for instance, motion vectors and object names.
Our tool supports two most widely-used 3D game engines: Unity and Unreal. The two engines are the backbone of the modern game industry and power almost every VR game\footnote{https://www.grandviewresearch.com/industry-analysis/game-engines-market-report}. 
In VR.net, 81\% of candidate games are implemented using the Unity engine, while the remaining ones are built on top of the Unreal engine. 
}

\subsubsection{Harnessing Graphics Labels from Unity}
Unity exposes C\# APIs for core rendering functions. As such, game developers can use high-level C\# scripts to implement game logic in a simpler and faster manner. 
We can also exploit those C\# interfaces to facilitate data collection.

\begin{enumerate} 
   \item \textbf{Camera Movement and FOV:} We first invoke a C\# API named \textit{`Camera.getAllCameras()'}, which returns an array of camera object references in the current scene. We then can access each object's \textit{worldToCameraMatrix} and \textit{projectMatrix} properties to obtain its view matrix and projection matrix.
   
   \item \textbf{Object Movement and Name:} We utilize an API named \textit{object.FindObjectOfType$<$Renderer$>$} to obtain a list of active \textit{renderer} instances in the current scene. A renderer instance corresponds to a 3D object visible on the screen. We can visit an instance's `localToWorldMatrix' property to access its model transformation matrix. We can also access its `name' and `bounds' members to capture the object's semantic name and 3D bounding box. 

      \item \textbf{DOF and Motion Vector:} We access a static member in the Camera class called \textit{`Camera.main'} to get an object reference of the active camera. We then update the camera's \textit{depthTextureMode} property to the combination of two flags \textit{DepthTextureMode.Depth} and \textit{DepthTextureMode.MotionVectors}. This instructs Unity to generate depth and motion vector textures for each frame internally. To access the texture contents, we first invoke a C\# API named \textit{Shader.GetGlobalTexture} with parameters `CameraDepthTexture' and `CameraMotionVectorsTexture'. They reveal native GPU resource pointers, which we then can copy to the system memory via the API `Graphics.copyTexture'.

\end{enumerate}
A challenging question is how to inject our data logging logic into a Unity game. 
The answer depends on the game's scripting backend.
A Unity game can use \textit{Mono}, an open-source C\# just-in-time compiler, to execute C\# scripts at runtime. In this case, we can extract C\# bytecode from the game folder (e.g., Mono/Managed) and conduct bytecode-level patching for code injection. An injection point is the `Camera.onPostRender' function, which is only called by the engine when it finishes rendering a frame. At this moment, every game scene data will be readily accessible.

Alternatively, the game may use IL2CPP, an ahead-of-time compiler, to translate C\# scripts into C source codes and compile them into native machine instructions before execution. This design complicates our tool's implementation since we can no longer inject high-level C\# code. 
Instead, we have to implement our data capture logic in shellcode, a sequence of machine instructions injectable into a game process's runtime memory with the intent to alter its behavior.
Compared to the C\# code, the shellcode must manually locate the aforementioned C\# wrappers' memory addresses. To achieve that, we can parse a metadata file named `global-metadata.dat' inside the game folder, which contains memory address offsets for all converted C\# functions. Alternatively, we can rely on an exported function called `il2cpp\_resolve\_icall' in the game process to query the wrappers' memory addresses at runtime.



\subsubsection{Capturing Graphics Labels from Unreal}
Unlike Unity, Unreal games are entirely implemented in C++. Moreover, they do not expose any engine interfaces to the outside world. 
Therefore, we need to take a different approach to data collection. The main intuition of our approach is to exploit the `reflection' mechanism in the Unreal engine.
Reflection is the ability of a program to query information about C++ classes and their member functions at runtime. 
To support this mechanism, Unreal internally maintains an array of metadata for all C++ objects in memory, called `GObject'. 
By examining this metadata via shellcode, our tool can locate important game scene objects and access their ground-truth data as follows. 
\begin{enumerate}
\item \textbf{Camera Movement and FOV:}  We initially filter out C++ objects with a class tag `ACameraActor' from the `GObject' array. This provides all camera objects' memory addresses. To identify active cameras, we can invoke a member function named `Engine.Actor.WasRecentlyRendered'. It returns true only if the camera is rendered in the last frame. Finally, we can utilize another member function called `Engine.CameraActor.GetCameraComponent' to retrieve a `ViewInfo' object, which encapsulates the camera's view and projection matrix. 

\item \textbf{Object Movement and Name:} We first separate objects with a class tag `AActor' from the `GObject' array. These objects correspond to every 3D object in the scene. To identify visible objects in the current frame, we can again use a member function `Engine.Actor.WasRecentlyRendered'. We can then extract its model transformation matrix and semantic names by invoking the member functions `GetTransform' and `GetActorNameOrLabel'.

\item \textbf{DOF and Motion Vector:} Unreal internally computes depth and motion vector texture to implement motion blur and temporal anti-aliasing effects. A core rendering function lies in a member function called `FDeferredShaderingSceneRender::RenderVelocities', which takes the native pointers to depth texture and motion vector texture as parameters. We can intercept the two parameters' contents by installing a hook on that function.
\end{enumerate} 

As a preliminary step, we need to locate the `GObject' memory address. We can reliably accomplish this task by conducting signature scanning through a game's runtime memory. Those signatures are universal (i.e., remain constant) for all Unreal games with the same engine version.
This is because Unreal's engine components are distributed as pre-built c++ object files. During the game development, those components are statically linked (i.e., combined as is) with each game's logic code to generate the final executable.

\subsubsection{Capturing Interaction Labels}
Our tool installs dynamic hooks on the low-level VR runtime \textit{OpenVR}\footnote{https://github.com/ValveSoftware/openvr} to query VR hardware status. Specifically, we intercept the functions \textit{IVRCompositor::WaitGetPoses} and \textit{IVRCompositor::GetControllerState} for headset movement and joystick interaction. 
So far, there is no standard interface to query gaze tracking and physiological signal.
Therefore, we must delegate the capture tasks to specific third-party applications from VR hardware vendors, for example, Tobii XR SDK\footnote{https://developer.tobii.com/tobii-pro-sdk/}.

To implement the verbal response self-report method, our tool conducts the voice recording with the help of the popular live streaming software \textit{OBS-studio}\footnote{https://github.com/obsproject/obs-studio/tree/master/libobs}.
We adopt the speech recognition toolkit \textit{Vosk}\footnote{https://alphacephei.com/vosk/} to automatically recognize users' voice responses. We utilize the largest voice model \textit{vosk-model-en-us-0.42-gigaspeech} in pursuit of high-accuracy recognition.

\subsection{Data Collection Procedure}
\textbf{Subjects:} Our data collection process was approved in advance by the Ethics Committee \edit{of the University of Auckland}.
Through email and posters, we first recruited 25 participants (14 Female, 11 Male, age range 23-40 years, mean = 27.4 years, standard derivation = 3.1).
They are students, faculty, and staff from three universities across three continents. These participants do not have eyesight issues and are familiar with VR. They were all provided with a participant information sheet before the data collection started.

\textbf{Equipment: }
We provide all participants with a pre-built VR setup,
consisting of an Alienware m15R6 gaming laptop and a Meta Quest Pro VR headset.
We pack all selected VR games into archive files and distribute them to participants using cloud storage. 
The archive files are in a portable self-extracting format to enable the simple `click and play' experience. 
Our data collection tool will automatically run as a background process when the gameplay starts. It requires no direct interaction from users.
\edit{
During the game sessions, we follow the FMS protocol and ask users to report their momentary self-reported scores verbally.} To encourage self-reporting, we flash a reminder message on the bottom right corner of the screen every 30 seconds. Participants could also voluntarily report whenever they felt a change in their discomfort level. 

\textbf{Procedure:}
Each participant was randomly assigned games from different genres. This assignment was done in a way that each game is played by at least five different users.
We instructed users to play every assigned game for around 15 minutes. Nevertheless, they were allowed stop participating for any reason, such as feeling overwhelmed. The data was collected in the participants' home settings without supervision. They could choose any convenient time slot to participate within two weeks. Once finished, they were instructed to upload their log file directories into cloud storage.

\section{Utilization of VR.net}~\label{sample}\edit{In this section, we showcase two essential utilities of VR.net. 
Firstly, VR.net can serve as a data set to construct risk factor detection systems for VR content. We provide two examples in section~\ref{exp1} and section~\ref{exp2}. The first one detects whether the camera is performing an excessive movement (i.e., fast speed, violent acceleration, and multi-axis rotation). 
The other example examines whether there is an excessive number of objects in the game scene (i.e., high object density). 
Recently, We are integrating these two systems into the Oculus Quest Store. Every time VR developers plan to publish a new game in the store, they are encouraged to upload a demo video. Our systems can automatically analyze the video to identify their risk factors. 
Secondly, VR.net can be used as a benchmark dataset to reproduce and validate results from previous motion sickness studies. 
In section~\ref{exp3}, we replicate a previous study that exploits the motion flow and depth texture labels to predict motion sickness levels. These examples underline the advantages of having a large-scale dataset with diverse VR games and rich labels. }

\subsection{Excessive Camera Movement Detection}~\label{exp1}
\begin{figure}[b]

  \begin{center}
  \includegraphics[width=0.65\textwidth]{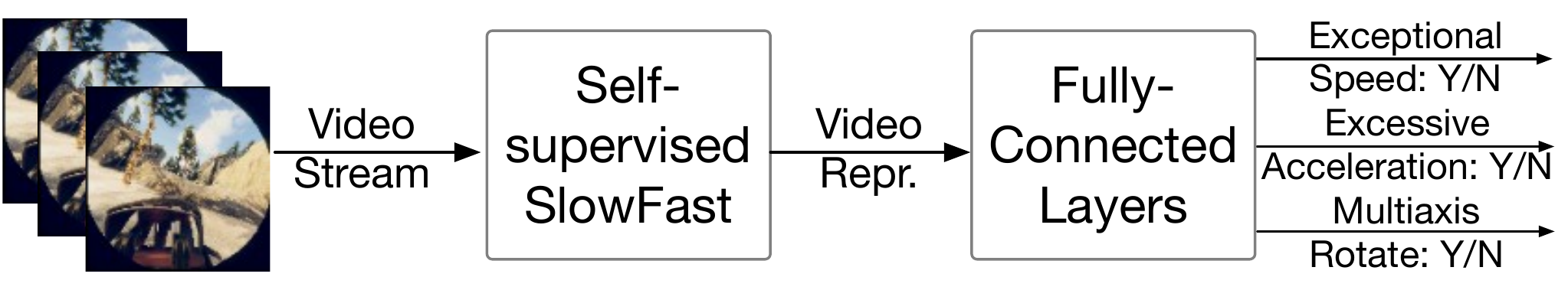}
  \caption{A pipeline to detect motion sickness risk factors.}
  \label{fig:exp1}
  \end{center}
  \end{figure}

We build a proof-of-concept system to predict \edit{camera-related} risk factors, as shown in Fig.~\ref{fig:exp1}. The model consumes a 1-second video stream and outputs three binary numbers, indicating the presence of the following risk factors:
\begin{enumerate}
 \item Fast camera speed: Whether the camera speed is faster than 3 m/s, at which most users will start experiencing VR sickness~\cite{so2001metric}.
 \item Excessive camera acceleration: Whether the camera acceleration is above a threshold of 0.7 m/$s^2$, at which participants start to feel temporary symptoms of cybersickness~\cite{terenzi2020rotational}.
\item Multi-axis rotation: Whether the camera executes a rotation movement involving more than one axis. Such rotation can significantly escalate the severity of discomfort~\cite{keshavarz2011axis}. More specifically, given two consecutive view matrices $V_1$ and $V_2$, we can calculate the camera rotation difference via the formula: $V_{\Delta} = V_2 * {V_1}^{-1}$, where $V_{\Delta}$ can be further decomposed into three elementary Euler angles, commonly referred to yaw, roll, and pitch. If two of them exceed a predefined value~\cite{bonato2009combined}, we detect a multi-axis rotation.
\end{enumerate}

Our model is built upon 
\textit{SlowFast}~\cite{feichtenhofer2019slowfast}, a commonly-used self-supervised learning model for video recognition. 
Self-supervised learning is a process where the model first trains itself to learn one part of the input from another part of the input. For example, given the upper half of the video frame, the model can learn how to predict the lower half of the frame. By doing this, the model can obtain useful representations of input videos, which can be later fine-tuned with a limited number of labels for actual supervised tasks. Self-supervised learning generally outperforms conventional supervised learning, particularly when the training dataset is small.

Since SlowFast already provides a model checkpoint pre-trained on sizeable video datasets, we could directly attach the model to fully-connected layers to output the risk factor labels. Straightforwardly, 
we could fine-tune three fully-connected layers, each of which outputs an individual label.
The merit is that it allows us to add more risk factors in the future without retraining the existing networks. Nevertheless, we opt for an alternative architecture, where we fine-tune a single fully-connected layer and output three labels together. This design exploits the fact that all the tasks may share the same underlying information. Training them together in one model can achieve higher performance (i.e.,  multi-task learning~\cite{fifty2021efficiently}).

We utilize the PyTorch~\cite{fan2021pytorchvideo} framework to implement the SlowFast model.
We select a roller coaster game in VR.net for model training. 
The game provides five roller coaster tracks on different terrains (e.g., snow mountain and urban city). Each track lasts approximately 6 minutes and is evaluated by ten different users. 
This generates a dataset with 5-hour video clips and the corresponding risk factor labels.
We first split the dataset into training and test sets in a ratio of 80/20. In our implementation, two splitting mechanisms are evaluated: 1) four tracks for training and one track for testing, and 2) four users for training and one user for testing.
We set the number of epochs to 5 and the batch size to 32. The learning rate is $1e^{-4}$. We use two V100 32 GB GPUs, and the training wall time is 12 hours.
The performance of our system for both data-splitting mechanisms is shown in Table~\ref{tab:my-table}.
Our system can predict three risk factors with reasonable accuracy regardless of the data-splitting mechanism. 
This also suggests that our model offers decent generalization performance across game content and users. 

\begin{table}[t]

\caption{Pipeline performance. Gray and Cray colors correspond to track-based splitting and user-based splitting, respectively.}

\label{tab:my-table}
\centering
\resizebox{0.50\textwidth}{!}{%
\begin{tabular}{|lll|}
\hline

\multicolumn{1}{|l|}{Risk Factors} & \multicolumn{1}{l|}{Accuracy} & F1 Score \\ \hline
\multicolumn{1}{|l|}{Excessive Speed} & \multicolumn{1}{l|}{\textcolor{gray}{0.76} / \textcolor{cyan}{0.82}} & \textcolor{gray}{0.71} / \textcolor{cyan}{0.74} \\ \hline
\multicolumn{1}{|l|}{Excessive Acceleration} & \multicolumn{1}{l|}{\textcolor{gray}{0.75} / \textcolor{cyan}{0.79}} & \textcolor{gray}{0.73} / \textcolor{cyan}{0.75} \\ \hline
\multicolumn{1}{|l|}{Multi-axis Rotation} & \multicolumn{1}{l|}{\textcolor{gray}{0.71} / \textcolor{cyan}{0.74}} & \textcolor{gray}{0.73} / \textcolor{cyan}{0.77} \\ \hline
\end{tabular}%
}

\end{table}

\subsection{Object Density Estimation}~\label{exp2}
Previous studies~\cite{kim2020deep} observed that a higher scene complexity is more likely to induce motion sickness.
For example, flight simulation gamers tend to perceive more severe sickness in a low-altitude flight than in a high-altitude one because the former scenario has more visual stimuli, such as buildings, trees, and cars.
Here, we exploit VR.net to build an object density estimator given a one-second gameplay video.

Specifically, We fine-tuned the state-of-the-art video recognition model VideoMAE~\cite{tong2022videomae} with the music game `Beat Saber' data in VR.net. The game was played by five users and each session lasted on average 3 minutes (i.e., 900 seconds in total).
We pre-processed the object movement data to exclude objects whose bounding boxes are too small. Those excluded objects are usually helper objects (e.g., light waves), which do not appear on the rendered frames. We then counted the distinct objects in one second and discretized the object numbers to generate object density classes: low (0-100 objects), medium (100-150 objects), and high ($>$150 objects). These thresholds are chosen to provide an approximately uniform data distribution across these three classes. 
Two sets of video data were used; In the first set, the videos were encoded at a frame rate of 30 FPS as usual. In the second set, the videos are resampled with 0.5X speed to increase the dataset size by a factor of 2.
These videos were then divided into the train, validation, and test sets in the proportion of 80/10/10. The 3-class classification accuracy of these models is provided in Table~\ref{tab:objectdensity}. 
This result shows that our simple way of exploiting
the object labels can provide high performance for the object density estimation task.
Our future investigations will include extending the models for more games.

\begin{table}[]
\centering
\caption{Object density estimation performance.}
\label{tab:objectdensity}
\resizebox{0.35\textwidth}{!}{%
\begin{tabular}{|l|l|l|}
\hline
FPS & Train/Val/Test & Accuracy \\ \hline
30 & 723 / 90 / 90 & 0.94 \\ \hline
15 & 1452 / 180 / 180 & 0.97 \\ \hline
\end{tabular}%
}
\end{table}
\vspace{-5pt}

\subsection{Comfort Rating Prediction}~\label{exp3}
\begin{figure}[t]
 \vspace{-5pt}
  \begin{center}
  \includegraphics[width=0.6\linewidth]{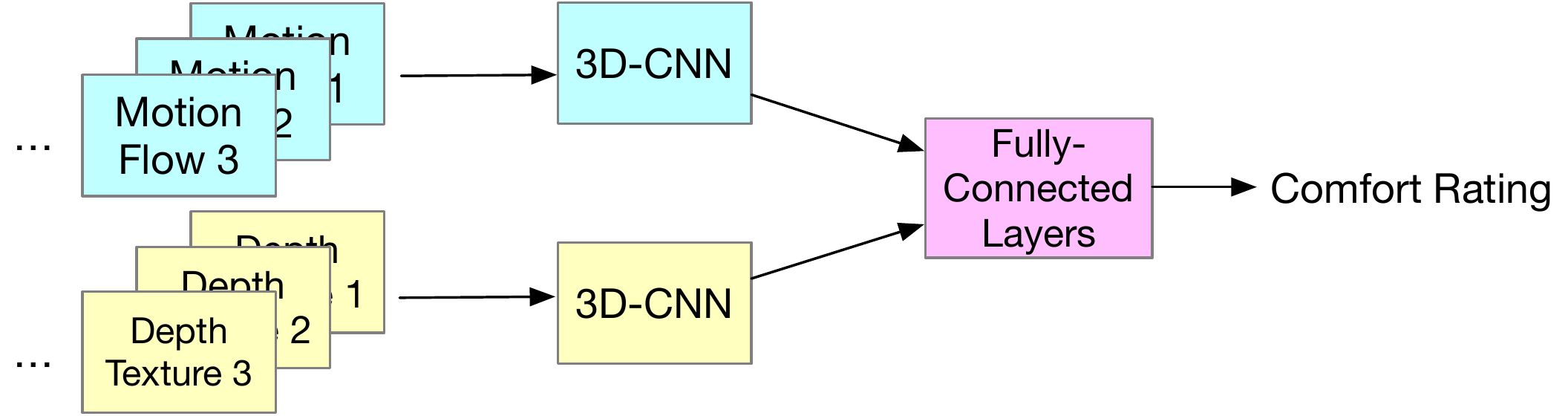}
 \vspace{-5pt}
  \caption{A 3D-CNN pipeline to predict sickness scores.}
  \label{fig:exp2}
  \end{center}
  \end{figure}

\edit{
In several previous studies~\cite{lee2019motion,padmanaban2018towards}, 
a convolutional neural network (CNN) has been deployed to predict the degree of motion sickness, as shown in Figure~
\ref{fig:exp2}.} The input features of the network include a sequence of motion flow frames and a sequence of depth texture frames. The output is a comfort rating on a scale from 1 to 5 (a higher number corresponds to a higher level of discomfort).
The CNN architecture is three-dimensional, which can perform convolution operations in the two-spatial dimensions (i.e., horizontal and vertical directions of a frame) and the temporal dimension (i.e., several consecutive frames). 
This enables expressing the degree of motion sickness with multiple frames as a single value. 
Despite these achievements,  these studies unanimously pointed out that the size of their training dataset limits their predictor performance.
Here, we can use VR.net as a benchmark dataset to replicate and validate this motion sickness research. 
 
We reuse the roller coaster game data in the first application. The video clips are encoded at a resolution of 1832$\times$960. We downscale each frame's size by $8$ (i.e., 229 $\times$ 120) to reduce computational cost. Amid the video stream, we discover 471 self-reported scores. We assign each score to a 3-second gameplay video clip right before the report. This generates a sub-dataset that contains 471 three-second video clips with their comfort ratings. The clip length is motivated by previous research~\cite{mchugh2019investigating}, which suggests that 3 seconds is sufficient for users to interpret VR content and generate a momentary cybersickness response. A deep-learning
model typically needs a considerable amount of training data. We thus use a simple-yet-effective pixel-shifting technique to augment our dataset.
Specifically, we shift the motion flow and depth texture by -5 and +5 pixels on horizontal and vertical axes with zero padding. Therefore, the amount of data increased to 4 times the original amount. Since it is challenging for human eyes to recognize the
differences between pixel-shifted frames, we can assume that the motion sickness induced by the shifted dataset is the same as that original dataset. 

We first split our dataset into training and test sets in a ratio of 90/10.
We then utilized the PyTorch framework to implement and train the 3D CNN architecture.
All experiments were performed on the same server used in the first application. We set the learning rate to $1e^{-4}$, and the total training time is approximately 14 hours.
Our results show that the model provides a decent prediction accuracy of 0.70.
Guided by the existing studies, we test the significance of the Pearson correlation coefficient between prediction and ground truth. We observed that the Pearson coefficient is strongly positive (0.74) with a p-value less than 0.05. This validates the previous finding that motion flow and depth of field play an essential role in sickness level prediction.

\section{Discussion and Future Work}
\textbf{Extending the size of the dataset}: 
We are continuing to expand the VR.net dataset. The next version of VR.net aims to offer 100-hour gameplay videos for 100 real-world VR games. 
To speed up the construction process, we plan to explore crowdsourcing marketplaces (e.g., Amazon Mechanical Turk) to recruit more participants. 
We also aim to deliver VR.net to wider research communities by making it more easily accessible. For instance, we are experimenting with InterPlanetary File System\footnote{https://ipfs.tech/} to enable the high-speed distribution of sizable video data in VR.net.
In addition, we attempt to foster a VR.net community by developing an online platform where everyone can contribute to and benefit from VR.net resources.  

\textbf{Building a full-fledged risk factor detection pipeline:}
We are exploring the potential of rich labels in VR.net.
Our primary goal is to increase the number of detectable risk factors in the pipeline in Fig.~\ref{fig:exp1}.
For instance, we can identify overly wide FOV, at which users are more susceptible to motion sickness~\cite{emoto2008viewing}. 
We can also detect the high magnitude of motion flow, which means the frame contains complex visual information and may increase the extent of the sickness~\cite{lee2017estimating}. 
We envision the full-fledged pipeline can be integrated into 3D design software such as Unreal and Unity editor. It can help content developers identify and mitigate risk factors in real-time.

\textbf{Exploiting VR.net for Generic Vision Tasks:}
We hope VR.net can also become a valuable resource for vision-related research. For instance, VR.net provides motion flow for each video frame. It thus can be used as benchmark datasets for existing optical flow algorithms. Our preliminary experiment compares two open-source algorithms, FlowNet 2~\cite{ilg2017flownet} and SPyNet~\cite{spynet2017}.
We apply their pre-trained models on each video frame and evaluate
the flow accuracy at a threshold of 3px for all valid pixels. Our results indicate that FlowNet2 delivers a moderately higher performance (11\%) than SPyNet, which seems to support the findings from a previous study~\cite{zhai2021optical}. 
Similarly, we also utilize the depth texture in VR.net to compare two depth estimation algorithms: Eigen et al. ~\cite{eigen2014depth} and Yin et al. ~\cite{yin2019enforcing}. These two algorithms are regression-based, meaning they take RGB frames as input and output depth textures. As a performance metric, we compute the root mean squared error between the output and the ground truth. 
Our results indicate that Yin et al. perform significantly better than their opponent. A potential reason is that it has a more complex network architecture and a higher number of network parameters.
In addition, our dataset provides every object's location and its bounding box. It thus can benefit object tracking and instance segmentation tasks.
In the future, we aim to evaluate more state-of-the-art vision algorithms and provide a comprehensive benchmark result.

\textbf{Limitations:}
VR games that do not build upon Unity and Unreal engines are excluded from our dataset. This is acceptable at this stage as they are scarce and do not significantly impact the large-scale and diverse properties of VR.net. 
In addition, our dataset does not include some less-known risk factors labels such as audio effects~\cite{kuiper2020knowing} and graphic realism~\cite{golding2012cognitive}. More research effort on them is warranted.

\section{Conclusion}
This paper presents `VR.net', a large-scale video dataset that aims to stimulate new ML-based VR sickness research and beyond. VR.net is sizable and covers a diverse range of real-world VR experiences.
In addition, VR.net provides rich label sets that are known to induce motion sickness. These ground truth labels are automatically and precisely extracted from 3D engines' rendering pipelines.
We showcase how VR.net can be used to build a machine-learning pipeline for risk factor detection and comfort rating estimation.
VR.net is constantly expanding. We envision its next version will offer 10X more data than the current form.

\bibliographystyle{ieee_fullname}
\bibliography{main}

\end{document}